\documentclass[letterpaper]{article} 
\usepackage{aaai23}  
\usepackage{times}  
\usepackage{helvet}  
\usepackage{courier}  
\usepackage[hyphens]{url}  
\usepackage{graphicx} 

\usepackage{amsmath} 
\usepackage{amssymb}
\usepackage{bm}
\usepackage{booktabs}
\usepackage{multirow}
\usepackage{tabularx}

\usepackage{natbib}  
\usepackage{caption} 
\usepackage{algorithm}
\usepackage{algorithmic}

\usepackage{newfloat}
\usepackage{listings}
\DeclareCaptionStyle{ruled}{labelfont=normalfont,labelsep=colon,strut=off} 
\floatstyle{ruled}
\newfloat{listing}{tb}{lst}{}
\floatname{listing}{Listing}
\title{ProSG: Using Prompt Synthetic Gradients to Alleviate Prompt Forgetting of RNN-like Language Models}

\author{
    Haotian Luo\textsuperscript\equalcontrib \quad Kunming Wu\equalcontrib \quad Cheng Dai\thanks{Corresponding Author}\\
    Sixian Ding \quad Xinhao Chen \\
}
\affiliations{
    \textsuperscript{\rm} Sichuan University\\

    haotianluo2002@gmail.com
}

\usepackage{bibentry}

\begin{document}
\maketitle
\begin{abstract}
RNN-like language models are getting renewed attention from NLP researchers in recent years and several models have made significant progress, which demonstrates performance comparable to traditional transformers. However, due to the recurrent nature of RNNs, this kind of language model can only store information in a set of fixed-length state vectors. As a consequence, they still suffer from forgetfulness though after a lot of improvements and optimizations, when given complex instructions or prompts. As the prompted generation is the main and most concerned function of LMs, solving the problem of forgetting in the process of generation is no wonder of vital importance. In this paper, focusing on easing the prompt forgetting during generation, we proposed an architecture to teach the model memorizing prompt during generation by synthetic gradient. To force the model to memorize the prompt, we derive the states that encode the prompt, then transform it into model parameter modification using low-rank gradient approximation, which hard-codes the prompt into model parameters temporarily. We construct a dataset for experiments, and the results have demonstrated the effectiveness of our method in solving the problem of forgetfulness in the process of prompted generation. We will release all the code upon acceptance.
\end{abstract}

\section{Introduction}
Transformer~\cite{vaswani2023attention} has long dominated the field of NLP across various domains due to its excellent performance and training parallelism, but the $\mathcal{O}(N)$ complexity of per step generation and history key-value pairs memory overhead makes transformers less efficient when the output sequence growing longer. As a consequence, researchers make a lot of efforts to explore a more efficient structure, for example, attention-free transformer(AFT)~\cite{zhai2021attention}. Lately, researchers have successfully developed powerful models that possess both parallel training and recursive inference capabilities, for example, RWKV~\cite{peng2023rwkv} and RetNet~\cite{sun2023retentive}.

Models like RWKV and RetNet combine the advantages of both transformers and RNNs, ensuring that the training complexity remains unchanged while achieving an inference process with a complexity of $\mathcal{O}(N)$. Through extensive experiments, this type of model surprisingly demonstrates performance comparable to traditional transformers.

However, it is also because of the recurrent nature of such models, which means compressing historical information into a set of fixed-length state hidden vectors, making it impossible for the model to retain all historical information. Therefore, it will no doubt lead to a series of forgetting problems. Take prompted generation as an example, given a prompt: “Write a story about Tom", the result may satisfy our expectations. But when the prompt is not a simple instruction but consists of two steps of operations, forgetfulness may occur. For example, given this prompt: “Write a story about Tom, this story should have a tragic ending” after the model generates relatively long content, it may forget the latter part of the given prompt, and not write a tragic ending as expected. We call this "prompt forgetting". We use this term to define the phenomenon where a language model, when given a specific prompt for generation, either forgets the specific details of the prompt or does not generate according to the requirements given by the prompt. Table 1 shows more cases of prompt forgetting.





\begin{table*}[!htb]
	\centering
	\begin{tabularx}{\textwidth}{lXXX}
		\toprule
		\textbf{Sample}&  \textbf{Prompt} & \textbf{Behaviour} & \textbf{Output}  \\
		\midrule
            
		A 
		& Write a story about student Noah, and write a summary for the story in the end.
		& Forget to summarize the story in the end of output.
		& Noah was a bright student who had always dreamed ... Eventually, he now works as a doctor at a hospital where he feels fulfilled with what he does every day.
		\\
		\midrule
		B 
		& List 10 character names, and write a story using these names.
		& Exactly list 10 names but forget to write the story.
		& 1. Sophia - A kind-hearted woman.
2. Jack - A charming man who loves to explore.
...
10. Max - A trustworthy man who is always there for his friends and family.\\
        \midrule
        C
        & Given a sequence of numbers: 1, 0, 1, 0, 0, 6, 2, 0, 0, 7, 2, 6, 0, 1, 2, 9. Now recite it.
        & Forget the given numbers.
        & The sequence is: 0, 1, 0, 1, 2, 0, 0, 1, 2, 6, 7, 2, 9
        \\
		\bottomrule
	\end{tabularx}%
	\caption{Examples for prompt forgetting}
 \label{tab:examples}
\end{table*}%

As enabling model obedience to the prompt is considered a paramount objective of the most important function of a language model~\cite{liu2021pretrain}, addressing the issue of prompt forgetting in the process of language model generation is undoubtedly imperative and calls for a resolution.

For the convenience of description and problem formulation, we have defined the term “multi-stage prompt” to refer to a prompt that consists of more than one sub-prompt or process (examples were shown in Table 1). Forgetting occurs because the length of the model state vector is fixed. During the process of inputting text into the model, there are situations where the model cannot determine whether to retain the information obtained at the current moment. As more text is inputted, important information also runs the risk of being forgotten. This problem is particularly detrimental for prompted generation tasks, which is the main function of large pre-trained language models.

\begin{figure}[h]
\hspace{-0.15in}
\includegraphics[width=3.85in]{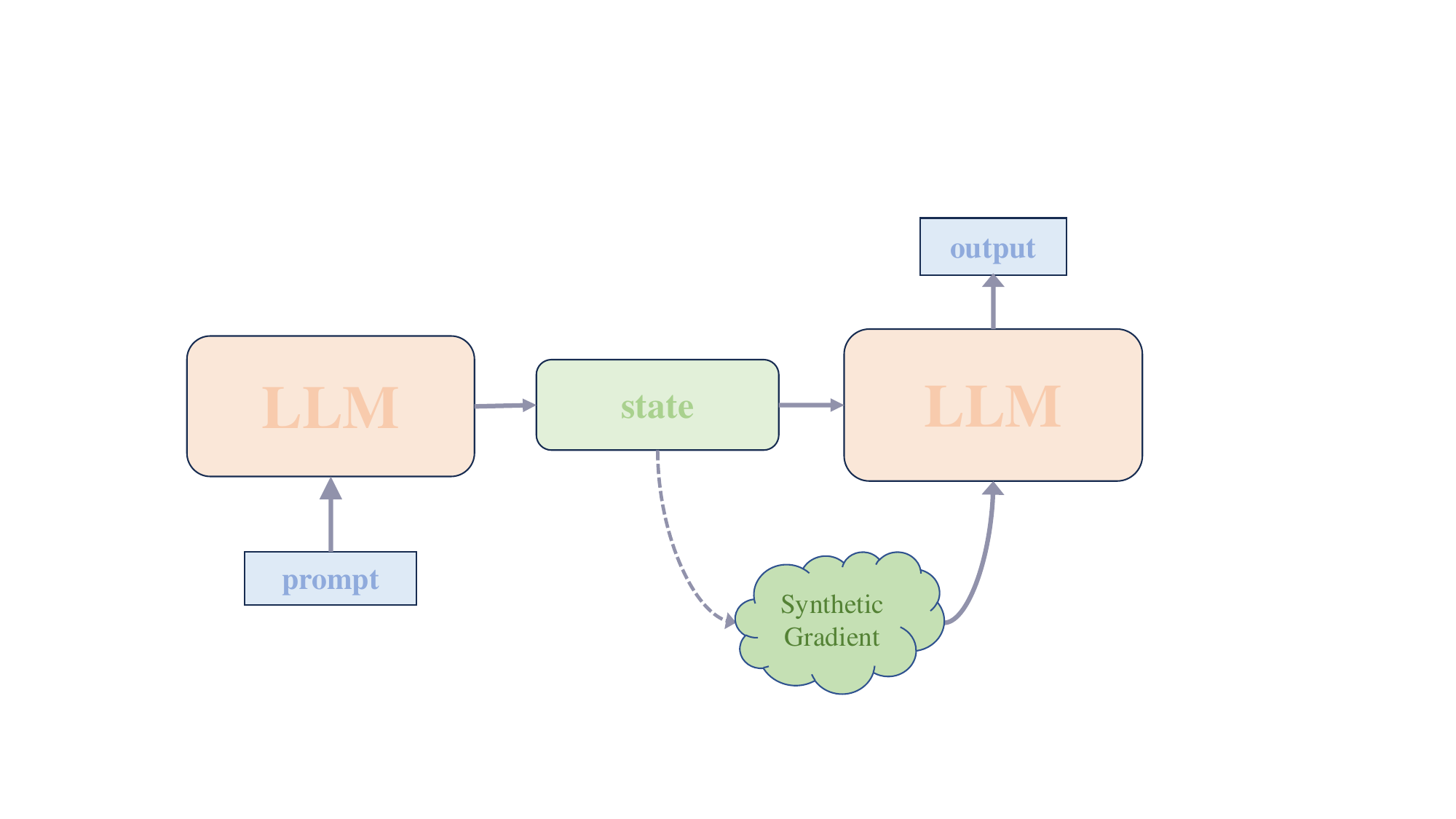}

\caption{Architecture Overview.}

\label{fig:model}
\end{figure}

 In this paper, aiming at alleviating the forgetfulness of prompt during generation, we exploit a structure called ProSG (Prompt Synthetic Gradient), that encodes prompt information into the model temporarily during one generation run to boost generation quality. First, in order to model the problem better, we build up a dataset, which contains 21k multi-stage prompts and corresponding answers following the common paradigm of instruction fine-tuning. Second, an adaptive module is utilized to calculate the gradients of the prompt and apply the change in the following generation. This adaptive module is a set of neural networks using low-rank adaption technique~\cite{hu2021lora} that take state which encode prompt as input and generate synthetic gradients~\cite{pmlr-v70-jaderberg17a} as output. The rough architecture is shown in Figure 1.

We conduct extensive and complete experiments to evaluate the effectiveness of our novel structure and dataset.  Experimental results show that our method leads to excellent generation performance in terms of both automatic metrics and human evaluation. In summary, our contributions are as follows: \\
(1)We evaluate the phenomenon of prompt forgetfulness and build a multi-stage instruction dataset(MuSI); \\ \\
(2) We propose a framework that facilitates better collaboration with datasets by temporarily encoding prompts into the model parameters, thus significantly alleviating the issue of prompt forgetting.

\section{Related Work}
\subsection{RNN-like LM}
After the efforts of researchers, there are currently two relatively successful RNN-like models, namely RWKV~\cite{peng2023rwkv} and RetNet~\cite{sun2023retentive}. This kind of model combines the advantages of both transformers and RNNs, ensuring that the training parallelism remains unchanged while achieving an inference process with linear complexity. Although these two models have decent performance, as RNNs, they are still relatively limited by fixed memory capacity, which is a direct consequence of fixed state vectors. This drawback leads to a noticeable forgetting effect, especially when giving the model multi-stage instruction for generation. The model forgets information about a given prompt as generation proceeds, or even before the generation has begun. Though the two models both claim to have a great performance, researchers of RWKV have released their code for a relatively long time. So we choose RWKV to conduct most of our research work.

\begin{figure*}[ht]
\centering
\includegraphics[width=6.65in]{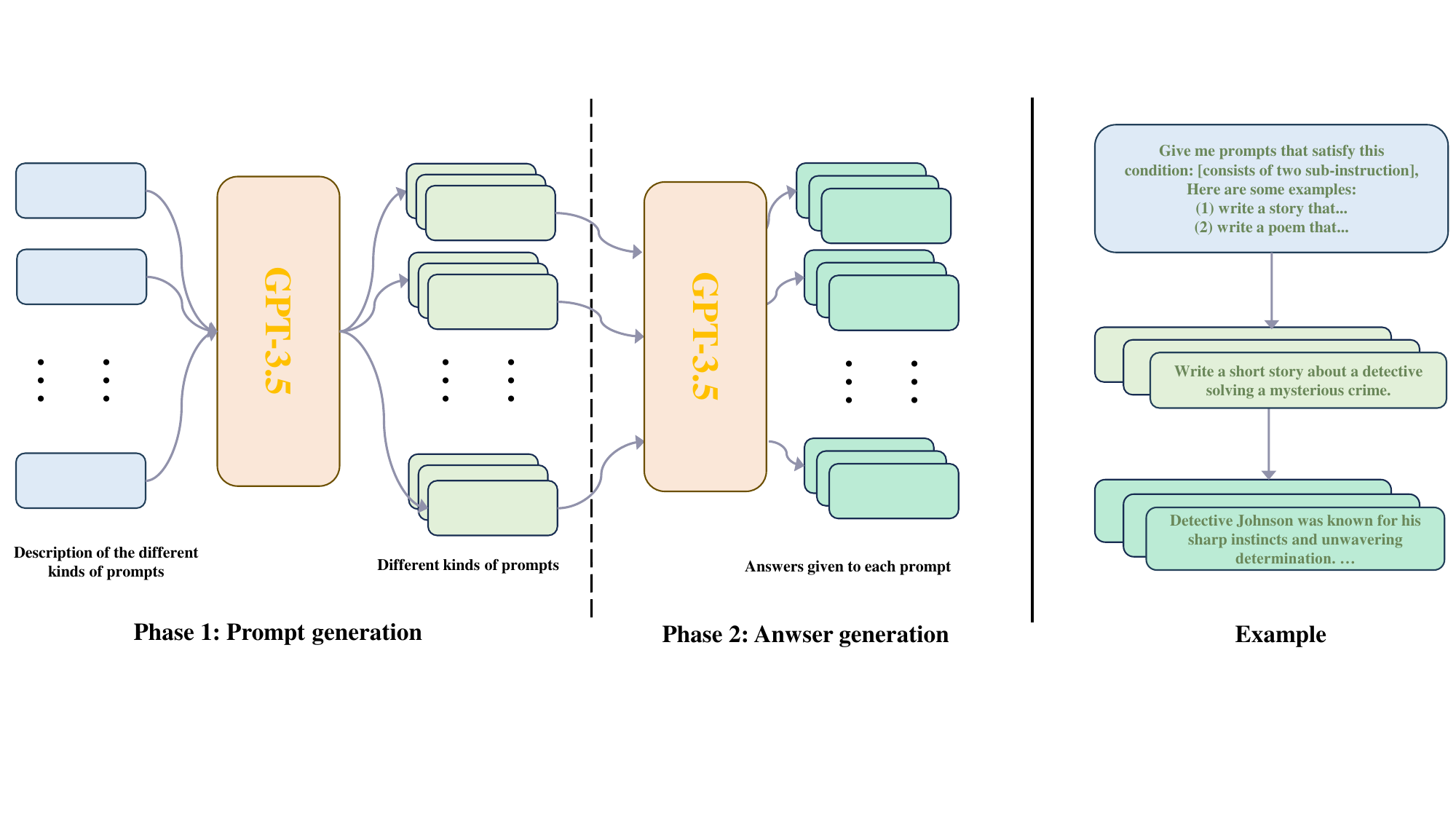}

\caption{Dataset Construction.}

\label{fig:dataset}
\end{figure*}

\subsection{ICL is Implicitly Fine-tuning}
ICL(in-context learning)~\cite{dong2023survey} is an emergent ability of large language models that enables them to adapt to various tasks when given a set of demonstration examples. This property has been demonstrated in LLMs like GPT-3 ~\cite{brown2020language}.
~\cite{dai2023gpt} provide a novel view of explaining ICL
as a process of meta-optimization, which considers LM as a meta-optimizer producing meta-gradients according to the demonstration examples through forward computation. Then gradients will be applied to the original LM through attention to build an ICL pattern.

Inspired by this view, we come up with the idea that consider instructed generation as a process of implicitly fine-tuning, which means that model can be seen to be optimized by the gradient from the prompt. So we can synthesize the gradient brought by the prompt to achieve the goal of mitigating forgetting.
\subsection{Synthetic Gradient}
Synthetic Gradients~\cite{pmlr-v70-jaderberg17a} is a method that is used for decoupling neural modules, aimed at accelerating the training process. In traditional neural network training, gradient information needs to be propagated from the output layer to the input layer through backpropagation~\cite{rumelhart1986learning}. Synthetic gradients, however, approximate the real gradient information by introducing an additional model. Usually, synthetic gradient model takes the intermediate layer outputs of the neural network as input and generates an approximate gradient, which is then passed to the previous layer of the network. In our framework, we utilize this technique to approximate gradients produced by prompt. 

\subsection{Low Rank Adaption}
LoRA~\cite{hu2021lora} is a technique to reduce the computational complexity and memory requirements when fine-tuning models. LoRA has been proven to deliver excellent results in various fine-tuning tasks in the field of NLP.

We make an assumption that the model's parameters need only minor modifications to achieve a certain level of memorization of prompt information which have the capability to enhance generation. So we use low-rank adaption when calculating prompt synthetic gradient to reduce the memory overhead of our framework.

\begin{figure*}[t]
\hspace{-5mm}
\includegraphics[width=7.7in]{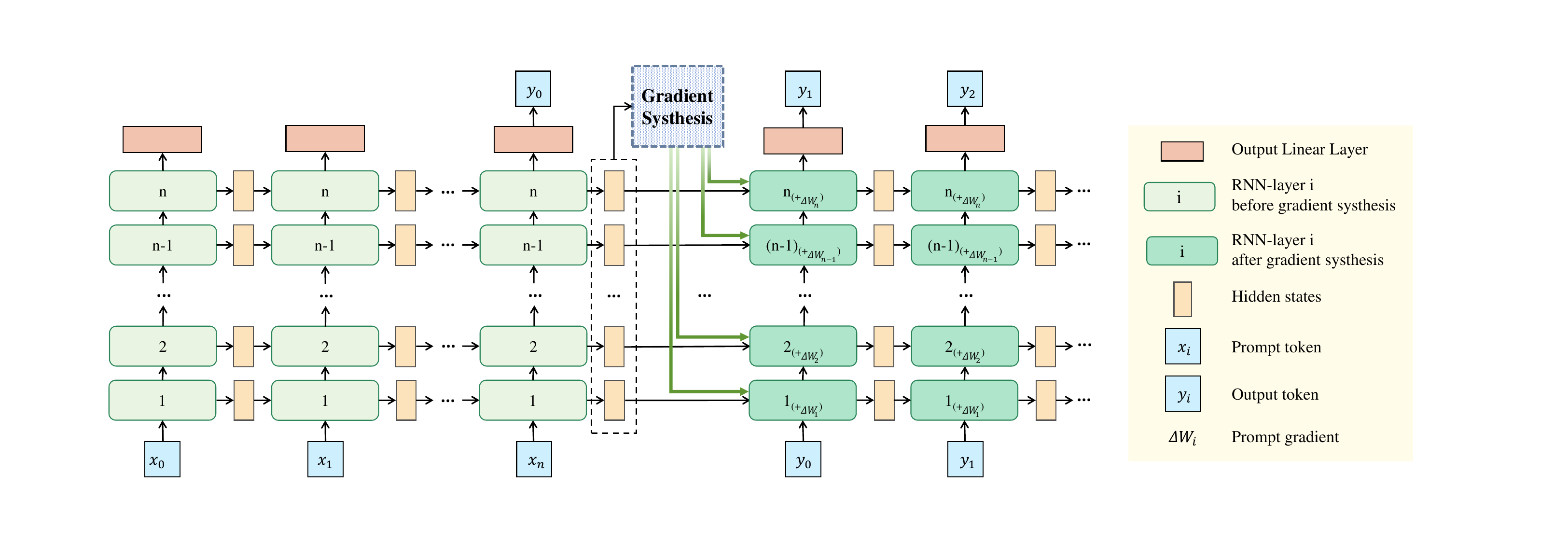}

\caption{Model architecture.}

\label{fig:architecture}
\end{figure*}

\section{Method}
\subsection{Dataset}
\subsubsection{Common Instruction vs. Multi-stage Instruction}
Most common instruction consists of only one operation/step, for example, "Write a story about Tom". The dataset used by Alpaca is a typical dataset containing mostly simple instructions. While multi-stage instruction consists of more than one operation/step, for example, "Write a story about Tom, and also make a summary at the end". For traditional language models using transformer architecture, the two kinds of instruction has slightly different, as the transformer can preserve complete information about instruction by storing K-V pairs. However, it is completely different for RNN-like models as the recurrent nature leads to an inevitable loss of information during generation, which disables RNN-like models from "looking back" the instruction like transformers. As a consequence, the forgetting phenomenon of the RNN-like model is particularly evident when dealing with multi-stage instructions.
\subsubsection{Multi-stage Instruction Dataset (MuSI)}
To deal with the forgetting process better, we leverage ChatGPT to collect a dataset named Multi-stage Instruction Dataset (MuSI), which contains a number of multi-stage instructions and corresponding answers, and then we manually eliminate non-compliant data. Finally, we collect 22k pieces of instruction-answer pair. It should be emphasized that the prompts selected for our dataset contain very little domain-specific knowledge, allowing for a fairer comparison with other large language models in public domains after fine-tuning. The whole construction process of MuSI is illustrated in Figure 2.

\subsubsection{Overview}
Our architecture consists of two parts, one of them is the backbone RNN-like LM, which is set to be RWKV as it has released most resources. Another part is prompt gradient synthesis module $\mathcal{G}$. For each generation, we first feed the prompt sequence $\mathbf{X}$, which contains n tokes $\{\mathbf{x_1,x_2,...,x_n}\}$ into RNN LM, then the hidden states vector $\mathbf{H_n}$ which containing prompt information will be utilized to generate synthetic gradient. Briefly, the hidden states will be used by gradient synthesis module as input, and output the low-rank approximate gradient  $\mathbf{\Delta W}$, which is expected to consistently convey prompt information in the process of generation by adding to the original parameter $\mathbf{W}$ as an increment. Thus in the follow-up generation, each forward pass 
that produces the $\mathbf{k}$-th output $\mathbf{y_k}$ token will be enhanced with the parameter increment $\mathbf{\Delta W}$. The whole process can be formulated as follows:

\begin{align}
\{ \mathbf{H_0,H_1...,H_n} \} &= \mathcal{RNN}(\mathbf{x_1,x_2,...,x_n}) \\
\mathbf{\Delta W} = \mathcal{G}(\mathbf{H^n}),\ & \mathbf{y_k} = \mathcal{RNN_{(+\mathbf{\Delta W})}}(\mathbf{y_{k-1}}) 
\end{align}

\begin{figure*}[ht]
\centering
\includegraphics[width=5in]{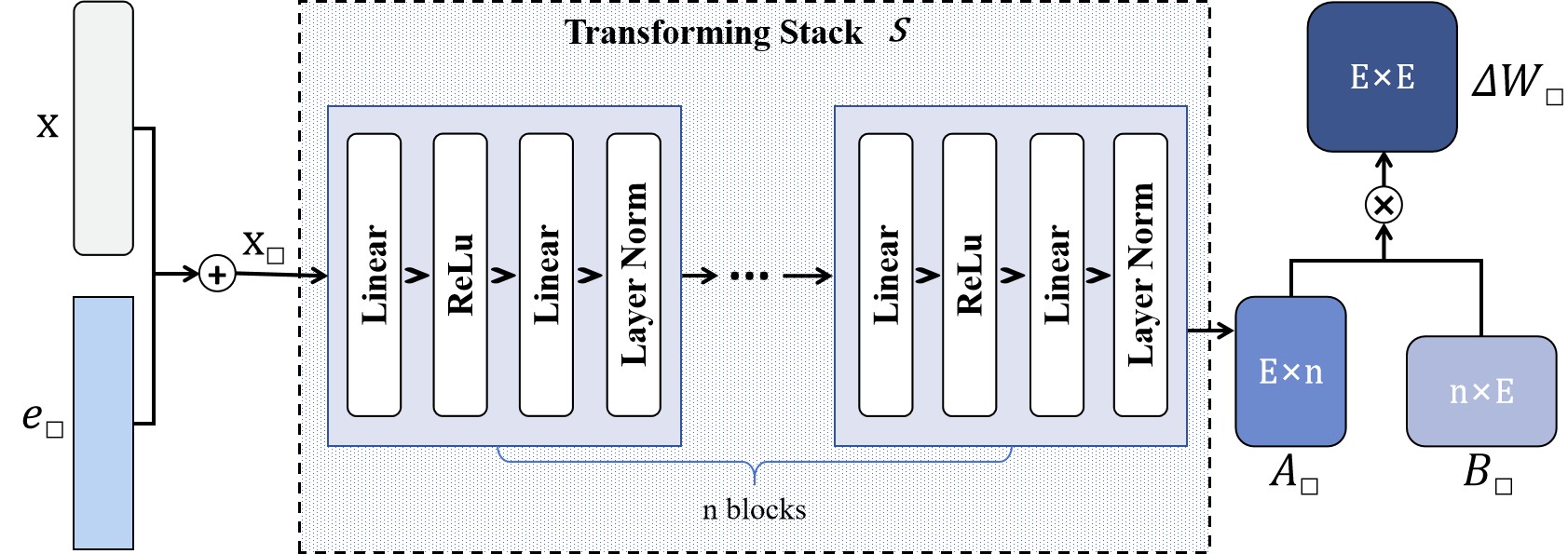}

\caption{Gradient synthesis module.}

\label{fig:gradient}
\end{figure*}

\subsubsection{RWKV}
RWKV is a RNN-like model that mainly comprised of a stack of residual blocks, each formed by a time-mixing sub-block and a channel-mixing sub-block with recurrent structure. In this paper, we append our gradient synthesis module to the time-mixing block.
The time-mixing block can be formulated as :
\begin{align}
\mathbf{r}_t &= \mathbf{W}_r \cdot (\mathbf{\mu}_r \mathbf{x}_t + (1 - \mathbf{\mu}_r) \mathbf{x}_{t-1} )\\
\mathbf{k}_t &= \mathbf{W}_k \cdot (\mathbf{\mu}_k \mathbf{x}_t + (1 - \mathbf{\mu}_k) \mathbf{x}_{t-1} )\\
\mathbf{v}_t &= \mathbf{W}_v \cdot (\mathbf{\mu}_v \mathbf{x}_t + (1 - \mathbf{\mu}_v) \mathbf{x}_{t-1} )\\
\mathbf{wkv}_t &= \frac{ \sum_{i=1}^{t-1} e^{-(t-1-i)\mathbf{w}+\mathbf{k}_i} \mathbf{v}_i + e^{\mathbf{u}+\mathbf{k}_t} \mathbf{v}_t}{\sum_{i=1}^{t-1} e^{-(t-1-i)\mathbf{w}+\mathbf{k}_i} + e^{\mathbf{u}+\mathbf{k}_t}}\\
\mathbf{o}_t &= \mathbf{W}_o \cdot (\sigma(\mathbf{r}_t) \odot \mathbf{wkv}_t)
\end{align}
The generation process will be modified by our synthetic gradient after computing prompt gradient. Thus time-mix computation will be:
\begin{align}
\mathbf{r}_t &= (\mathbf{W}_r+\Delta \mathbf{W}_{r}) \cdot (\mathbf{\mu}_r \mathbf{x}_t + (1 - \mathbf{\mu}_r) \mathbf{x}_{t-1} )\\
\mathbf{k}_t &= (\mathbf{W}_k+\Delta \mathbf{W}_{k}) \cdot (\mathbf{\mu}_k \mathbf{x}_t + (1 - \mathbf{\mu}_k) \mathbf{x}_{t-1} )\\
\mathbf{v}_t &= (\mathbf{W}_v+\Delta \mathbf{W}_{v}) \cdot (\mathbf{\mu}_v \mathbf{x}_t + (1 - \mathbf{\mu}_v) \mathbf{x}_{t-1} )\\
\mathbf{wkv}_t &= \frac{ \sum_{i=1}^{t-1} e^{-(t-1-i)\mathbf{w}+\mathbf{k}_i} \mathbf{v}_i + e^{\mathbf{u}+\mathbf{k}_t} \mathbf{v}_t}{\sum_{i=1}^{t-1} e^{-(t-1-i)\mathbf{w}+\mathbf{k}_i} + e^{\mathbf{u}+\mathbf{k}_t}}\\
\mathbf{o}_t &= (\mathbf{W}_o+\Delta \mathbf{W}_{o}) \cdot (\sigma(\mathbf{r}_t) \odot \mathbf{wkv}_t)
\end{align}
The computation of other parts in RWKV model leave unchanged.
\subsubsection{State Embedding}
Due to the distinct feature spaces of the matrices for the states key, value, receptance, and output, we used the same gradient synthesis module to approximate the gradients of the four parameters in the same layer. This approach helped reduce the parameter count and enable parallel computation. This method speeds up computation, but it also has a fatal flaw: different parameters should receive different updates when calculating prompt synthetic gradient. To overcome this issue, we introduced state embedding.

State embedding is a set of learned embedding vectors for each different state that store in each layer, which will be added to the input $\mathbf{x_{\square}}$ before being fed to the core module (transforming stack $\mathcal{S}$) of gradient synthesis module $\mathcal{G}$. 

State embedding enables the use of the same module to process inputs in four different modes, allowing a single module to provide synthetic gradients with distinct properties for the four kinds of state.

\subsubsection{Transforming Stack}
Transforming stack $\mathcal{S}$ is a set of identical blocks, which is a sequential model which contains linear layers, activation function, and layer norm. In detail, a transforming block consists of a linear layer, a ReLU activation function, another linear layer, and a layer norm. We also add a residual connection between the output and input. For a single block. The transforming stack is the core module used for gradient synthesis.

\subsubsection{Output Low-rank Matrix}
The shape of the state vector is $1\times\mathbf{E}$, and the shape of the state matrix is $\mathbf{E}\times\mathbf{E}$. It should be noticed that we augment the state to be a vector of shape $1\times\mathbf{nE}$, where $\mathbf{n}$ is chosen from 1, 2 or 3. The state vector remains unchanged after being processed by the transforming stack and then will be reshaped to $\mathbf{n}\times\mathbf{E}$. Based on the assumption that prompt memorization can be achieved to some extent by low-order changes in the model parameter space, we employed the technical of low-rank adaption. In order to transform the output shape to $\mathbf{E}\times\mathbf{E}$, we design a learned output low-rank matrix $\mathbf{B}$ of shape $\mathbf{E}\times \mathbf{n}$, which will multiple with the output of transforming stack $\mathcal{S}$.

\subsubsection{Gradient Synthesis Module}
Our gradient synthesis module consists of a transforming stack $\mathcal{S}$, a set of state embedding $\mathbf{e}$, and a set of output low-rank matrix $\mathbf{B}$. This module is shown in Figure 4.
Since the detailed design of hidden state varies with the model architecture, our formulation is specified with RWKV in order to show module details more clearly. The state of each layer will be fed into Gradient Synthesis Module $\mathcal{G}$ and calculate gradients. Every layer performs the same operation, in order to formulate the whole process in detail, we select the $\mathbf{i}$-th layer as an example.
In our experiments, we choose the input $\mathbf{x}$ of each layer, which encodes prompt to add with 4 different state embeddings:
\begin{align}
\mathbf{x_k} = \mathbf{x} + \mathbf{e_k} \\
\mathbf{x_v} = \mathbf{x} + \mathbf{e_v} \\
\mathbf{x_r} = \mathbf{x} + \mathbf{e_r} \\
\mathbf{x_o} = \mathbf{x} + \mathbf{e_o} \\
\end{align}
where k, v, r, o stand for key, value, receptance and output respectively. And then each $\mathbf{x_{\square}}$ will be processed by the transforming stack and produce
$\mathbf{A_{\square}}$:
\begin{align} 
\mathbf{A_{\square}} = \mathcal{S}(\mathbf{x_{\square}}) \\
\end{align}
After this, low-rank matrix $\mathbf{A_{\square}}$ will multiply with corresponding 
$\mathbf{B_{\square}}$ and produces $\mathbf{\Delta W_{\square}}$.
\begin{align}
\mathbf{\Delta W_{\square}} = \mathbf{B_{\square}} \times \mathbf{A_{\square}}\\
\end{align}


\begin{table*}[h]
\centering
\begin{center}
    \begin{tabular}{l@{\hspace{1.4cm}} c@{\hspace{1.4cm}} c@{\hspace{1.4cm}} c@{\hspace{1.4cm}} c@{\hspace{1.4cm}} c}
    \toprule
    \multirow{2}{*}{Models} & \multirow{2}{*}{Perplexity} & \multirow{2}{*}{Accuracy} & \multicolumn{3}{c}{Human Evaluation}  \\
   \cmidrule(lr){4-6}
     & & & Fluency & Accuracy & Quality \\
    \midrule
    \noalign{\smallskip}
    Vicuna-7B & \textbf{--} & 0.760 & \textbf{--} & \textbf{--} & \textbf{--}\\
    \noalign{\smallskip}
    ChatGLM-6B & \textbf{--} & \textbf{0.938} & \textbf{--} & \textbf{--} & \textbf{--}\\
    \noalign{\smallskip}
    GPT-2-0.4B (MuSI) & \textbf{--} & 0.893 & \textbf{--} & \textbf{--} & \textbf{--}\\
    \noalign{\smallskip}
    \midrule
    \noalign{\smallskip}
    RWKV-4-0.4B & 5.480 & 0.534 & 4.03 & 3.42 & 3.45\\
    \noalign{\smallskip}
    RWKV-4-0.4B (MuSI) & 3.583 & 0.698 & 4.10 & 3.57 & 3.58\\
    \noalign{\smallskip}
    RWKV-4-0.4B (MuSI, ProSG)  & \textbf{3.161}  &\textbf{ 0.761} & \textbf{4.22} & \textbf{3.87} & \textbf{3.78} \\
    \bottomrule
    \end{tabular}

\end{center}
\caption{Experiment results. PPL of first three language models is not computed. Because they each has different tokenization, the results are not comparable.}
\label{tab:results}
\end{table*}

\subsection{Training Strategy}
\subsubsection{Parallel Training}
During training, the gradient synthesis module requires access to the hidden states of the encoded prompt $\mathbf{X}$, specifically the final state vector. To achieve this, we initially pad the several prompts $\{\mathbf{X_1,X_2,...,X_k}\}$, which has a length of $\{\mathbf{l_1,l_2,...,l_k}\}$ and input a batch into the model, thus obtaining the state vectors sequence $\{\mathbf{H_1,H_2,...,H_k}\}$, $\mathbf{H_i}\in \mathbb{R}^{\mathbf{L} \times \mathbf{E}}$, where $\mathbf{L}$ is the padding length and $\mathbf{E}$ is channel dimension. However, due to varying prompt lengths and the introduction of padding tokens, the information in the vector at the end of the prompt might be lost. The required is in fact to be $\{\mathbf{H_1^{l_1},H_2^{l_2},...,H_k^{l_k}}\}$,$\mathbf{H_i^{j}}\in \mathbb{R}^{\mathbf{E}}$ instead of $\{\mathbf{H_1^{L},H_2^{L},...,H_k^{L}}\}$. Therefore, during processing, it's essential to pass the original length of the prompt. This allows us to use the vector at the position corresponding to the original length of the prompt as the required state vector and also ensures consistency between training and inference.

\subsubsection{Two-Phase Training}
We used a two-stage training approach, where the first stage involved fine-tuning the backbone language model, which a training loss, and the second stage focused on training the gradient synthesis module to further enhance its memory capability. Formally, let the dataset be $\mathcal{D}$ with size $N$. the first stage aims to maximize the log-likelihood (denoted by $\mathcal{J}^{RNN}$) of the target over all the training samples of $\mathcal{D}$, that is,
\begin{align}
\mathcal{J}^{RNN}& = \sum_n^N \sum_t^{T_n} \log P_{\theta_1}(x_{n,t}|\bm x_{n,<t})
\end{align}
where  $x_{n,t}$ stands for the $t$-th word of the $n$-th sample. $T_n$ denotes the word length of the sample $\bm y_n$. $\theta_1$ is the model's parameters.
The second stage also aims to maximize the log-likelihood of all training samples, but the parameter of backbone model $\theta_1$ is frozen, which can be formulated as follows:
\begin{align}
\mathcal{J}^{S}& = \sum_n^N \sum_t^{T_n} \log P_{\theta_1,\theta_2}(x_{n,t}|\bm x_{n,<t})
\end{align}
In this formula, $\mathcal{J}^{S}$ denotes the log-likelihood of the second stage, which optimizes the gradient synthesis module, which has parameter $\theta_2$
\section{Experiment}
\subsection{Dataset}
Due to the lack of a suitable dataset that can effectively meet the requirements of multi-stage instructions at present, our experiments are all conducted on the MuSI dataset. 
\subsection{Competing Methods/Models}
We will compare the model with models that has a similar size, and also widely recognized models with much more parameters which is capable of generating satisfying outputs. We describe several chosen models as follows:

\textbf{GPT-2, Vicuna, ChatGLM}
GPT-2~\cite{Radford2019LanguageMA}, Vicuna~\cite{vicuna2023}, ChatGML ~\cite{zeng2022glm} are all pre-trained decoder-only transformers. They have the capability to comprehend the context of input text and generate coherent and logical output text. As transformers, they retain the entire history token sequence. Therefore, we consider these three models as benchmarks.

\textbf{RWKV}
For RWKV, We designed three experiments: (1) Original fin-tuned version of RWKV-4-World; (2) MuSI fine-tuned; (3) MuSI fine-tuned plus ProSG architecture. This experimental setup allows us to analyze whether MuSI and ProSG each play a role in mitigating prompt forgetting.

\subsection{Evaluation Metircs}
We adopted three analytical metrics: Perplexity, Accuracy, and Human Evaluation.

\textbf{Perplexity}
Perplexity, or PPL, is a quantitative measure that indicates how well a language model can predict a given set of words based on its training data. Lower perplexity values suggest that the model is better at making accurate predictions.

The perplexity of language models is the exponential of cross-entropy and is used to indicate the predictive capability of the model. Assuming we have a set of test data denoted as $D$, consisting of $N$ samples, each represented as $x_i$, where $x_i$ is a tokenized textual sequence. The model predicts the probability distribution $p_i(y)$ for the next token given each sequence $x_i$, where $y$ represents the next token.

For each sample $x_i$, the formula to compute cross-entropy is:
\[
H(x_i) = -\frac{1}{M} \sum_{j=1}^M \log p_i(y_j)
\]
Here, $M$ is the number of tokens in sample $x_i$, $y_j$ is the $j$-th token in the sample, and $p_i(y_j)$ is the predicted probability of the $j$-th token by the model.
Then, sum up the cross-entropy for all samples to obtain the cross-entropy for the entire dataset:
\[
H(D) = \sum_{i=1}^N H(x_i)
\]
Finally, perplexity (PPL) can be calculated using cross-entropy:
\[
PPL(D) = e^{H(D)}
\]

\begin{table*}[!htb]
	\centering
	\begin{tabularx}{\textwidth}{X}
		\toprule
            \textbf{Instruction} \newline What are three common types of exercise routines? Briefly describe each one and recommend one for someone looking to improve their cardiovascular fitness.\\
		\midrule
            \textbf{RWKV-4-0.4B (MuSI, ProSG)} \newline 
            Three common types of exercise routines are: \newline
1. Cardiovascular exercise: This type ... \newline
2. Strength training: This type ... \newline
3. Yoga: Yoga is a ... \newline
\textbf{For someone looking to improve cardiovascular fitness, I would recommend a combination of aerobic exercises and strength training exercises in order to improve cardiovascular fitness and overall health.}\\
		\midrule
            \textbf{RWKV-4-0.4B (MuSI)} \newline
            Three common types of exercise routines are cardio, strength training, and flexibility exercises.\newline
1. Cardiovascular Exercise: This involves ...\newline
2. Strength Training: This involves ...\newline
3. Flexibility Exercise: This involves ... \\
            \midrule
            \textbf{RWKV-4-0.4B} \newline
            1. Walking: Walking is a low-impact exercise that ... \newline
2. Cycling: Cycling is a low-impact exercise that ...\newline
3. Swimming: Swimming is a low-impact exercise that ...\\
		\bottomrule
	\end{tabularx}%
	\label{tab:case}%
	\caption{Case study}
\end{table*}%

\textbf{Accuracy}
We define accuracy as whether the generated content follows the given prompt. If an output answers all the questions or considers all the sub-instructions in the given prompt, this output will receive one point; otherwise, it will receive a zero. 

This evaluation method has the following two advantages:(1)allows us to focus on analyzing whether the model's generation follows the prompt, thereby determining the degree of forgetting during generation. (2) by not focusing on the model's learning of the dataset's language style during fine-tuning, we can compare it with other models that have not been fine-tuned by MuSI.

However, manually analyzing a large volume of model outputs for compliance is quite challenging and time-consuming. Therefore, we leverage ChatGPT with appropriate prompts to achieve the goal of a rough automated analysis. Through our validation, this approach has demonstrated a relatively reliable accuracy as a rough automatic evaluation.

\textbf{Human Evaluation}
As we observe that ChatGPT fails to make correct judgments on certain questions, such as those involving mathematical calculations, and also for the purpose of achieving a more comprehensive and granular analysis for three RWKV variations, we conducted manual evaluations. We asked five annotators to analyze and compare the quality, fluency, and accuracy of the generated results in a blind fashion. Scoring ranges of 0 to 5 were established for all three indicators. 

These three indicators have the following definition :\\
\textbf{1) Fluency}: to measure the fluency of generated sentences and identify any occurrences of repetitive generation.\\
\textbf{2) Accuracy}: same as that mentioned in the last subsection.\\
\textbf{3) Quality}: to assess whether the provided answers are logically coherent, appropriately detailed, and aligned with human preferences.\\

\subsection{Results}
\subsubsection{Automatic Evaluation Results}
Table 2 presents the performance of all the models. From the table, we can observe that RWKV, after fine-tuning with the MuSI dataset, achieved a higher accuracy score. This indicates that our dataset indeed mitigated the model's forgetting of prompts. Furtherly, the use of the ProSG architecture enhanced the model's memory capacity, which can be inferred from the improvement in indicators. Our methods made the model's performance much closer to three transformers, which have complete memory capabilities. This exactly aligns with our expectations.

\subsubsection{Human Evaluation Results}
The results of human evaluation are reported in the right part of Table 2. We can see that, our dataset and architecture both have contributed to the improvement of the model's performance. The results are consistent with the automatic metric.
\subsubsection{Case Study}
In order to have an intuitive presentation of the generated results, we select a few prompts and record the output of several different models, which is shown in Table 3. Due to space constraints, we only show 1 sample and display the key parts of the results, omitting many less significant details. The complete results can be found in the supplementary materials. We can observe a significant improvement in the model's output after incorporating the ProSG architecture, achieving the desired effect of prompt memorization. This once again demonstrates the effectiveness of our framework.

\section{Conclusion}
In this paper, we specifically analyze the prompt forgetting problem during the process of generation. We construct a multi-stage instruction dataset (MuSI) and propose a gradient synthesis-based architecture(ProSG) to alleviate prompt forgetting. Extensive experiments demonstrate that our approach successfully enhances the model's memory capacity for prompts, enabling better generation tasks under prompt control.

\section{Limitation}
Although our method has achieved promising results, due to resource limitations, we were unable to conduct related experiments on larger language models such as 13B or 33B. Future research should be undertaken to explore the prompt forgetting and the corresponding solution of large language models. 

\bibliography{aaai23}

\end{document}